\documentclass[conference]{IEEEtran}

\IEEEoverridecommandlockouts
\usepackage{cite}
\usepackage{amsmath,amssymb,amsfonts,adjustbox}
\usepackage{algorithmic}
\usepackage{graphicx}
\usepackage{textcomp}
\usepackage{xcolor}
\usepackage{booktabs}
\usepackage[utf8]{inputenc}
\def\BibTeX{{\rm B\kern-.05em{\sc i\kern-.025em b}\kern-.08em
    T\kern-.1667em\lower.7ex\hbox{E}\kern-.125emX}}
    
\makeatletter
\newcommand{\linebreakand}{%
  \end{@IEEEauthorhalign}
  \hfill\mbox{}\par
  \mbox{}\hfill\begin{@IEEEauthorhalign}
}
\makeatother
    
\begin{document}
\thispagestyle{empty}
\onecolumn 
\vspace*{\fill}
This work has been accepted at the IEEE Symposium Series On Computational Intelligence 2022. Copyright may be transferred without notice, after which this version may no longer be accessible.
\vspace*{\fill}

\twocolumn
\newpage 

\title{Fully Complex-valued Fully Convolutional Multi-feature Fusion
Network ($FC^2$MFN) for Building Segmentation of InSAR images}

\author{
\IEEEauthorblockN{ Aniruddh Sikdar $^ {\dagger 1}$  \thanks{$\dagger$ \  \textnormal{Equal contribution of authors.}}}
\and
\IEEEauthorblockN{ Sumanth Udupa $^ {\dagger 2}$}
\and
\IEEEauthorblockN{Suresh Sundaram $^{2}$}
\linebreakand \IEEEauthorblockN{Narasimhan Sundararajan$^{2}$}
\linebreakand \\
$^{1}$ {Robert Bosch Centre for Cyber-Physical Systems, Indian Institute of Science,
Bengaluru, India.}\\
$^{2}$ {Department of Aerospace Engineering, Indian Institute of Science,
Bengaluru, India.}
}

\maketitle

\begin{abstract}
Building segmentation in high-resolution InSAR images is a challenging task that can be useful for large-scale surveillance. Although complex-valued deep learning networks perform better than their real-valued counterparts for complex-valued SAR data, phase information is not retained throughout
the network, which causes a loss of information. 
This paper proposes a Fully Complex-valued, Fully Convolutional
Multi-feature Fusion Network($FC^2$MFN) for building semantic segmentation on InSAR images using a novel, fully complex-valued learning scheme. 
$FC^2$MFN learns multi-scale features, performs multi-feature fusion, and has a complex-valued output.
For the particularity of complex-valued InSAR data, a new complex-valued pooling layer
is proposed that compares complex numbers considering
their magnitude and phase. This helps the network retain
the phase information even through the pooling layer. Experimental results on the simulated InSAR dataset
\cite{2csm-3723-20} show that $FC^2$MFN achieves better results compared to
other state-of-the-art methods in terms of segmentation
performance and model complexity.
\end{abstract}


\begin{figure*}[ht]
    \centering
    \includegraphics[width=0.95\linewidth]{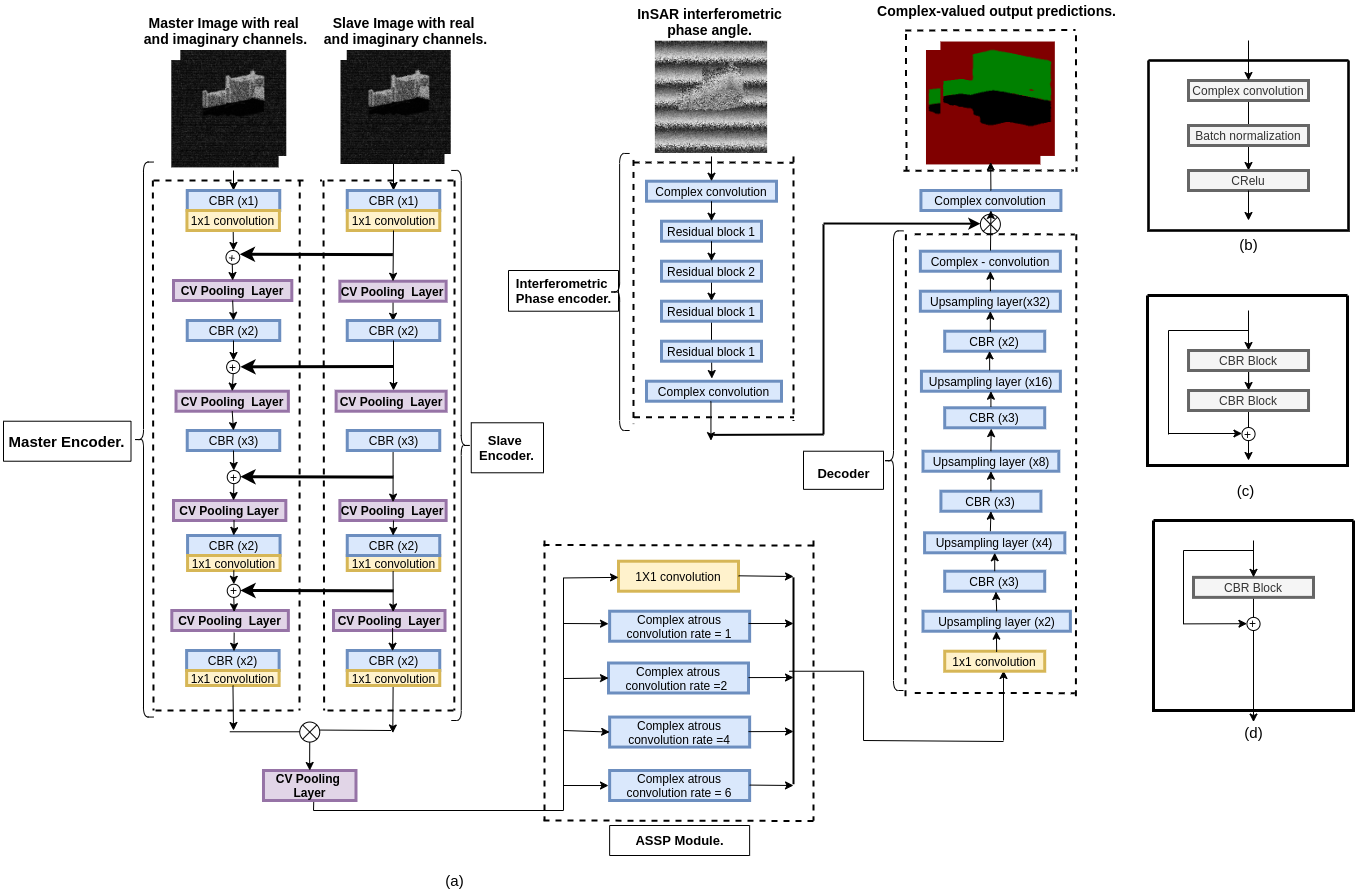}
    \caption{(a)Network architecture of$FC^2$MFN with complex-valued(CV) pooling layers. (b) CBR blocks- Complex-valued convolution layer followed by batch normalization and CRelu. The filter size in CBR blocks are 3x3. (c)Residual block 1. (d)Residual block 2. The output of $FC^2$MFN is a complex-valued feature map with real and imaginary channels.}
    \label{fig:one-hot encoding}
\end{figure*}

\section{Introduction}
\label{sec:intro}
Deep learning has made tremendous progress in computer vision-related tasks due to its robust feature representation capability.
Semantic segmentation and object detection are helpful for large-scale satellite image surveillance tasks.
Using EO sensors for an extended period is not feasible for such applications as they cannot capture important information in all weather and low light conditions.
Synthetic Aperture Radar(SAR) images can be used to map the terrain as they can retain information in all weather conditions.
The performance of Automatic Target Recognition(ATR) systems can significantly benefit from using SAR images both InSAR and PolSAR, for monitoring the changes in the terrain.
Change detection of buildings on the terrain has applications like 3-D modeling, map updating, and urban change monitoring\cite{chen2021cvcmff}.\\
SAR images are complex-valued data with amplitude and phase components. The amplitude component contains information about the wave that is bounced back to the radar and depends on the reflectance property of the terrain. The interferometric phase component of an InSAR image contains the height information of the targets and buildings on the terrain\cite{chen2021cvcmff} \cite{RAUCOULES2007289}.
In high-resolution SAR images, scattering properties like shadow, layover, and single scattering carry useful characteristics about the buildings and are reflected in both the amplitude and the interferometric phase of the InSAR image\cite{chen2021cvcmff}.
Semantic segmentation architectures like Unet\cite{DBLP:journals/corr/RonnebergerFB15}, PSPNet\cite{zhao2017pyramid}, RefineNet\cite{lin2017refinenet} and Deeplav3+\cite{DBLP:journals/corr/abs-1802-02611} have achieved competitive results on optical semantic segmentation benchmark datasets like NYUDv2\cite{Silberman:ECCV12}, PASCAL VOC 2012\cite{Everingham15} and Cityspaces\cite{cordts2015cityscapes}.
These networks are termed real-valued networks as they have real-valued weights, real-valued activation functions and
real-valued input.
Projecting complex-valued InSAR data to
the real domain by using only the amplitude as input and
discarding the phase information is not an optimal way of
learning as the phase contains valuable information regarding the terrain and should be used by these state-of-the-art
networks for building segmentation tasks.\\
Complex-valued deep learning approaches are more effective when dealing with complex-valued data\cite{6789140}.
Chen \textit{et al.}\cite{9397870} introduced CVCMFFNet for building semantic segmentation of InSAR images on simulated InSAR building dataset\cite{2csm-3723-20}.
CVCMFFNet takes complex-valued InSAR images as input and performs multi-scale and multi-feature fusion. Magnitude operation converts the complex-valued output to the real domain before the final softmax operation, and predictions are made using a real-valued output feature map.
The main drawback of projecting the complex-valued data in the real domain is the loss of phase information which is crucial for image reconstruction of complex-valued SAR data.\\
In this paper, we present a new complex-valued learning scheme for Fully Complex-valued Fully Convolutional Multi-feature Fusion Network ($FC^2$MFN) for semantic segmentation on the simulated InSAR dataset\cite{2csm-3723-20}.  
Overall, the main contributions of this paper can be summarized as follows:\\
(1)A novel fully complex-valued learning scheme is proposed for $FC^2$MFN to operate and learn in the complex domain  to avoid the loss of phase information using orthogonal decision boundary theory.\\
(2) A new complex-valued pooling layer is proposed to compare two complex numbers using the magnitude and the phase information.\\
(3) $FC^2$MFN network outperforms other state-of-the-art networks in segmentation performance on the simulated InSAR dataset\cite{2csm-3723-20} and is computationally efficient compared to other networks in teams of floating-point operations(FLOPs).

\section{Related work}

\subsection{Real-valued semantic segmentation}

Convolutional networks have achieved the state of the art results for dense prediction tasks.
Fully convolutional network(FCN) proposed by Shelhamer  \textit{et al.} \cite{long2015fully} is an end-to-end model built to perform segmentation for arbitrary size of the input and map output to the input resolution.
A novel skip architecture is proposed with in-network upsampling and multi-layer skip connections to combine fine and coarse layers. The skip connections refine the spatial precision of the output.
UNet proposed by Ronneberger \textit{et al.}\cite{DBLP:journals/corr/RonnebergerFB15} is used for segmenting medical images in an end-to-end manner from less number of data samples.
It is a U-shaped autoencoder with a contracting path to capture context and an expanding path for precise localization.
Chen \textit{et al.}\cite{DBLP:journals/corr/abs-1802-02611} proposed Deeplab v3+ for dense prediction task.
It incorporates Atrous Spatial Pyramid Pooling (ASPP) block to capture multi-scale information. It uses depthwise separable convolution in the ASSP block and the decoder to make the network faster.
SegNet introduced by Badrinarayanan \textit{et al.}\cite{badrinarayanan2017segnet} is used for scene understanding applications as it is efficient in terms of memory and inference time.
It has a symmetric encoder-decoder structure with a decoder corresponding to each encoder.
The decoder performs non-linear upsampling by using the pooling indices stored while performing the max-pooling operation in the encoder.

\subsection{Complex-valued semantic segmentation for SAR}
Complex-valued deep learning models are used to perform segmentation tasks when dealing with complex-valued SAR data.
Yu \textit{et al.}\cite{yu2022lightweight} proposed a lightweight complex-valued DeepLabv3+ for semantic segmentation of PolSAR image to avoid overfitting. Magnitude operation is used to project the complex-valued data to the real domain, and predictions are made using a real-valued feature map.
Cao \textit{et al.} \cite{cao2019pixel} used a fully complex-valued network for PolSAR image classification.The softmax function is applied to the output complex-valued feature map to calculate the prediction probability map. Cross-entropy loss function is used to compare the complex-valued predictions with the ground truth labels.

\section{Methodology}
In this section, complex-valued convolution operation\cite{DBLP:journals/corr/TrabelsiBSSSMRB17} is explained followed by the newly proposed complex-valued pooling layer and $FC^2$MFN. 
Finally, the fully complex-valued learning scheme using orthogonal decision boundary theory and complex-valued loss is explained.

\subsection{Fully Complex-valued Fully Convolutional Multi-feature Fusion Network($FC^2$MFN)}
\subsubsection{Complex-valued convolution operation} Similar to real-valued 2D convolutional operator, the complex-valued convolution operator is defined for complex-valued weight W =$W_R +\emph{i}W_I$ and complex-valued input I=$I_R +\emph{i}I_I$ where $W_R,W_I,I_R$ and $I_I$ are real-valued entities\cite{DBLP:journals/corr/TrabelsiBSSSMRB17}.Complex valued convolution is defined as follows,
\begin{equation}
  \begin{aligned}
    \textbf{W} \ast \textbf{I} & = ( W_R +\emph{i}W_I) \ast ( I_R +\emph{i}I_I) \\ 
  \end{aligned}
\end{equation}
where $W_R$ ,$W_I$ denote the real and the imaginary parts of the complex-valued weight, and $I_R$ ,$I_I$ denote the real and imaginary parts of the complex-valued input.  

\subsubsection{Complex-valued Pooling layers}
The pooling layers commonly used in complex-valued deep learning focus only on the amplitude of the complex numbers and discard the phase.
To avoid this loss of information, a new pooling layer is proposed that compares complex numbers by considering their magnitude and phase information.
\begin{figure}[ht]
    \centering
    \includegraphics[width=8.75cm]{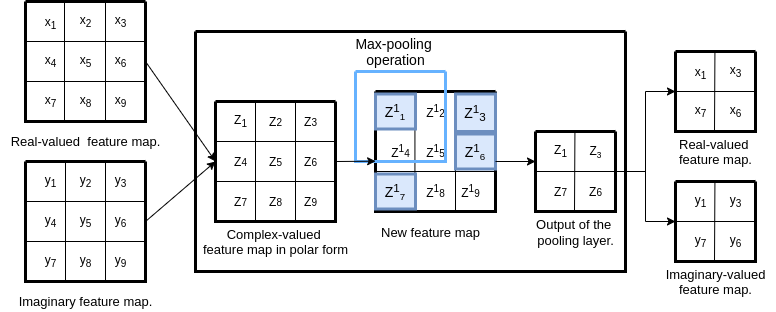}
    \caption{ shows the pooling operation. The complex-valued feature map is the input to the pooling layer. New feature map is constructed using (4). Max-pooling operation is performed on the new feature map, and the output is downsampled. The indices are recorded for upsampling layers.}
    \label{fig:max_pool}
\end{figure}
Let \textbf{C} and \textbf{R} denote the complex-valued domain and the real-valued domain respectively. Complex numbers $\emph{z}_1$ and $\emph{z}_2$ are represented as,\\
\begin{equation}
  \begin{aligned}
    z_1 & = x_1 + \emph{i}y_1 ,  \quad    \text{where} \quad  x_1,y_1 \in \textbf{R} \\ 
    z_2 & = x_2 + \emph{i}y_2 ,  \quad    \text{where} \quad  x_2,y_2 \in \textbf{R}
  \end{aligned}
\end{equation}
The complex-valued feature maps are converted from cartesian form to the polar form,
\begin{equation}
  \begin{aligned}
    z & = |r| e ^ {i\theta} 
  \end{aligned}
\end{equation}
where  r = $\sqrt{x^2 + y^2}$  and $\theta=\arctan $ $ (\frac{y}{x})$.\\
The complex-numbers $z_1$ and $z_2$ are converted to $z^1_1$ and $z^1_2$ using the following equations,\footnote{Detailed derivation for the pooling layer can be found in the supplementary material.}\\
\begin{equation}
z_1^1 = \left\{
        \begin{array}{ll}
            |r_1|^2 + \frac{1}{|r_1|^2} + 2 \cdot cos(2 \theta_1), \quad |r_1| > \delta \\
            |r_1|^2  + 2 \cdot cos(2 \theta_1), \quad \quad \quad \text{otherwise.}
        \end{array}
    \right.
\end{equation}
\begin{equation*}
z_2^1 = \left\{
        \begin{array}{ll}
            |r_2|^2 + \frac{1}{|r_2|^2} + 2 \cdot cos(2 \theta_2), \quad |r_2| > \delta \\
            |r_2|^2  + 2 \cdot cos(2 \theta_2), \quad \quad \quad  \text{otherwise.}
        \end{array}
    \right.
\end{equation*}
where $\delta$ can be tuned manually and is used for numerical stability.
The max-pooling operation computes the maximum of $z_1$ and $z_2$ as follows:
\begin{equation}
max(z_1,z_2) = \left\{
        \begin{array}{ll}
           z_1 & \text{if} \quad z^1_1 > z^1_2 \\
           z_2 & \text{if} \quad  z^1_1 < z^1_2
        \end{array}
    \right.
\end{equation}
The input to the pooling layer is the complex-valued feature map in cartesian form. The feature map is first converted to polar form as shown in Fig.\ref{fig:max_pool}.
Each element of the feature map is updated using (4), and a new feature map is constructed. Max-pooling operation is performed on this new feature map. The recorded indices are used to downsample the real and imaginary feature maps.
\subsubsection{Network architecture}
$FC^2$MFN is an extension of CVCMFFNet\cite{chen2021cvcmff}.
It has an encoder-decoder structure connected by the  Atrous Spatial Pyramid Pooling (ASPP) block\cite{DBLP:journals/corr/abs-1802-02611}, as shown in Fig.\ref{fig:one-hot encoding}(a).
The network has three inputs - master and slave SAR image and their interferometric phase angle. The master and slave SAR images are two channeled inputs with real and imaginary components. The interferometric phase angle is a single channel real-valued input.\\
The input data is forward propagated through the encoders. Complex-valued convolutional layers and pooling layers are used to downsample the feature maps.
The master and slave encoders follow a similar structure of VGG-16\cite{Simonyan15}, but without some of the convolutional layers, as shown in Fig.\ref{fig:one-hot encoding}(a).
The phase encoder is different from the master and slave encoder and has complex-valued residual blocks, as shown in Fig.\ref{fig:one-hot encoding}(c) and Fig.\ref{fig:one-hot encoding}(d).
All components of $FC^2$MFN  are complex-valued components.
The CBR blocks consist of three layers - complex valued convolution layer, followed by batch normalization and CRelu\cite{DBLP:journals/corr/TrabelsiBSSSMRB17}, as shown in Fig.\ref{fig:one-hot encoding}(b).
As the master and slave images are strongly correlated, their features are fused in the master channel of the encoder.
The features of the interferometric phase angle are fused with the output of the decoder as the phase features are not strongly correlated with the master and slave features\cite{9397870}.
Since pooling layers are computationally expensive, pointwise convolutional layers\cite{lin2014network} are introduced in the master and slave encoders.
The downsampled feature maps from the encoder are the input to the ASSP block.
The ASSP block assists the network in countering the multi-scale issue.
It contains four atrous separable convolutional layers with different dilation rates and one 1x1 convolutional layer. The output of all layers are concatenated and fed to the decoder.
The decoder block used in $FC^2$MFN is similar to the one proposed in CVCMFFNet\cite{9397870}, but it is fully complex-valued.
The low-resolution feature maps are upsampled to higher-resolution feature maps in the decoder using the pooling indices of the complex-valued pooling recorded in the encoder.
The output is a complex-valued feature map with real and imaginary channels.

\subsection{Fully complex-valued learning}
\begin{figure}[ht]
    \centering
    \includegraphics[width=8.5cm]{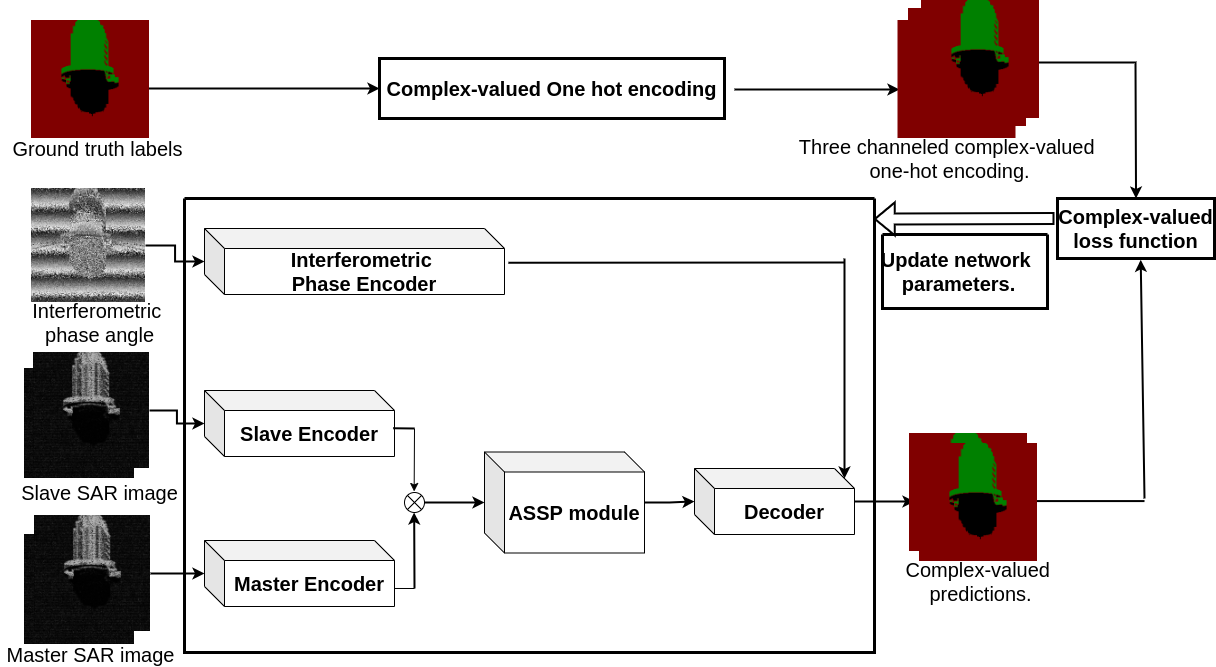}
    \caption{Block diagram of fully complex-valued training scheme. The labels are converted to complex-domain via one-hot encoding. The output predictions of $FC^2$MFN is compared with the labels using complex-valued loss function.}
    \label{fig:block_diagram}
\end{figure}
\subsubsection{Complex-valued one hot encoding}
To compare with the complex-valued predictions, the real-valued ground truth labels are one-hot encoded to complex-valued labels, as shown in Fig.\ref{fig:block_diagram}.
Let $\{(x_1,c_1),(x_2,c_2)....(x_t,c_t) ... (x_n,c_n) \}$ be the data samples, where $x_t$ represents the t-th input data sample and $c_t$ represents its corresponding label\cite{Suresh2013}.
For the label  $c_t$ of the t-th sample, one-hot encoding $y_k^t$ =$\{y_1^t,y_2^t,..y_k^t...y_n^t\}$ is as follows:
\begin{equation*}
y_k^t = \left\{
        \begin{array}{ll}
            1 + i, & \text{if} \quad c_t=k \\
            -1 -i, & \quad \text{otherwise.}
        \end{array}
    \right.
\end{equation*}

\subsubsection{Orthogonal boundary theory}Decision boundary of a fully complex-valued neural network consists of two hypersurfaces - real and imaginary\cite{6789140}. If the activation function used in the network satisfies the Cauchy–Riemann equations, then these hypersurfaces are orthogonal to each other (lemma 6.1 \cite{Suresh2013},\cite{6414644}).
The real and imaginary hypersurfaces of a network with CRelu activation function are orthogonal to each other as Crelu satisfies Cauchy–Riemann equations\cite{DBLP:journals/corr/TrabelsiBSSSMRB17}.
The ground truth labels are one hot encoded to the complex-domain in the output feature space.
The real or imaginary hypersurface is sufficient to classify the on and off values of the one-hot encoded labels and train the network.
Using orthogonal decision boundary theory, the ground truth labels can be directly compared with the real-channel of the complex-valued output feature map to compute accuracy and other task-specific performance metrics.

\subsubsection{Complex-valued loss function}
A new loss function is proposed to train $FC^2$MFN in the complex-domain\cite{Suresh2013}.
The loss function computes the error between one hot encoded labels and the predictions. This helps the phase information be preserved and considered when updating the network parameters.
The complex-valued loss \textbf{e} is given by ,
\begin{equation}
e = \left\{
        \begin{array}{ll}
            
            (y_l^t)-(\hat y_l^t), & \label{eq1}
        \end{array} 
    \right.
\end{equation}
where $(y_l^t)$ is the complex-valued one hot encoding and $(\hat y_l^t)$ is the complex-valued predictions.\\
Complex-valued cost functions are not used as loss functions due to lack of ordering in the complex-domain\cite{5236983}.
Real-valued loss function E is defined as the product of complex-valued loss \textbf{e} and its complex conjugate\cite{Suresh2013-2},
\begin{equation}
  \begin{aligned}
      E & =\frac{1}{2n} (e^H)(e),  \label{eq2}\\
  \end{aligned}
\end{equation}
where n is the number of data points. The labels are one hot encoded into a three-channeled feature map as there are three ground truth labels in the simulated InSAR dataset\cite{2csm-3723-20}. The output of $FC^2$MFN is also a three-channeled complex-valued feature map. The proposed loss function is used to compare the two and train the network.

\begin{table*}[]
\centering
\caption{Comparison of class-wise segmentation performance of $FC^2$MFN with state of the art models.}
\begin{tabular}{|c|c|c|c|c|}
\hline
Method     & IoU of shadow($\%$) & IoU of ground($\%$) & IoU of layover($\%$) & Mean IOU($\%$) \\ \hline
UNet       & 89.74            & 92.57            & 88.47             & 90.26       \\ 
SegNet     & 76.50            & 81.30            & 66.99             & 74.93       \\ 
RefineNet  & 89.69            & 94.34            & 88.35             & 90.79       \\ 
PSPNet     & 81.72            & 91.97            & 80.09             & 84.60       \\ 
Deeplabv3+ & 94.01            & 96.43            & 92.86             & 94.43       \\ 
CVCMFFNet  & 94.76            & 97.24            & 96.12             & 96.04       \\ \hline
$FC^2$MFN     & \textbf{98.074}           & \textbf{99.44}            & \textbf{98.01}             & \textbf{98.508}      \\ \hline
\end{tabular}%
\label{tab:1}
\end{table*}

\begin{table*}[]
\caption{Ablation study of fully complex-valued $FC^2$MFN with real valued(RV) and complex-valued(CV) networks.}
\centering
 
\begin{tabular}{|c|c|c|c|c|c|c|c|}
\hline
Method &
  OA(\%) &
  MPA(\%) &
  Test loss &
  IoU of shadow(\%) &
  IoU of  ground(\%) &
  IoU of  layover(\%) &
  \begin{tabular}[c]{@{}c@{}}Mean \\ IoU(\%)\end{tabular} \\ \hline
SegNet         & 88.3 & 77.6 & 0.355 & 76.5 & 81.3 & 67.0 & 74.9 \\
SegNet+ASSP    & 91.1 & 81.5 & 0.318 & 87.1 & 76.5 & 71.3 & 78.3 \\
CV-SegNet      & 96.6 & 94.1 & 0.208 & 91.3 & 95.1 & 86.3 & 90.9 \\
CV-SegNet+ASSP & 97.1 & 95.3 & 0.203 & 92.5 & 95.8 & 85.8 & 91.4 \\
CV-DeepLabv3+  & 98.3 & 97.2 & 0.177 & 94.2 & 97.7 & 95.2 & 95.7 \\
CVCMFFNet      & 98.5 & 97.9 & 0.169 & 94.8 & 97.2 & 96.1 & 96.0 \\ \hline
$FC^2$MFN &
  \textbf{99.48} &
  \textbf{99.39} &
  \textbf{0.0550} &
  \textbf{98.074} &
  \textbf{99.44} &
  \textbf{98.01} &
  \textbf{98.508} \\ \hline
\end{tabular}%

\label{tab:3}
\end{table*}

\section{Performance evaluation of $FC^2$MFN}
In this section, the justifications for using $FC^2$MFN for semantic segmentation on the InSAR dataset\cite{2csm-3723-20},\cite{chen2021cvcmff} are presented.
Before proceeding with the results, the details about the InSAR dataset are first described in section 4.1.
In section 4.2, the effectiveness of the new fully complex-valued learning scheme is shown by evaluating the segmentation performance with other state of the art methods like UNet\cite{DBLP:journals/corr/RonnebergerFB15}, SegNet\cite{badrinarayanan2016segnet}, Deeplab v3+\cite{DBLP:journals/corr/abs-1802-02611} and CVCMFFNet\cite{9397870}.
$FC^2$MFN is trained on the NVIDIA RTX 2080 Ti GPU for 100 epochs (10k iterations) with a batch size of 2.
The network is implemented in Tensorflow and trained using the Adam optimizer\cite{DBLP:journals/corr/KingmaB14} with a learning rate of 1e-5.
\subsection{Dataset description}
\begin{figure}[ht]
    \centering
    \includegraphics[width=8.5cm]{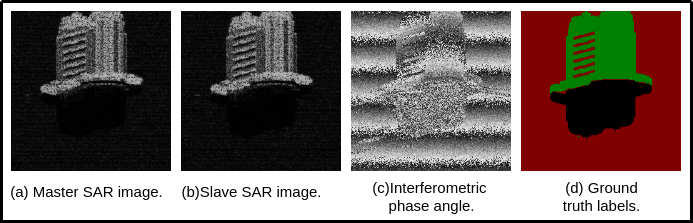}
    \caption{Data sample from the InSAR dataset. Each sample has a master and slave image along with the corresponding interferometric phase angle. The ground truth labels with their corresponding colors as shown in Table  \ref{tab:2}.}
    \label{fig:dataset}
\end{figure}

\begin{table}
\caption{Ground truth labels and their color mark. }
  \centering
  \begin{tabular}{|c|c|c|}
    \hline \
    Labels & Scattering property & Color mark \\ \hline
    \midrule
    0 & Shadow & Black \\
    1 & Ground & Red \\
    2 & Layover & Green \\ \hline
    \bottomrule
  \end{tabular}
  
  \label{tab:2}
\end{table}
The dataset used in this study is the high-resolution synthetically generated InSAR dataset \cite{2csm-3723-20} \cite{9397870}.
When building segmentation is performed using SAR, the scattering phenomena like layover, single scattering, and shadow are important characteristics.
The dataset is synthetically generated using these characteristics.
Table \ref{tab:2} shows the ground truth labels and their corresponding color codes.
Fig.\ref{fig:dataset} shows a data sample from the simulated InSAR building dataset. Each data sample contains a master and slave SAR image, their corresponding interferometric phase, and the ground truth label.
The dataset has 312 SAR image pairs. The training set has 216 data samples, and the test set has 96 data samples.
The size of the image is 256x256.
\subsection{Semantic segmentation results}
\begin{figure*}[ht]
    \centering
    \includegraphics[width=1.0\linewidth]{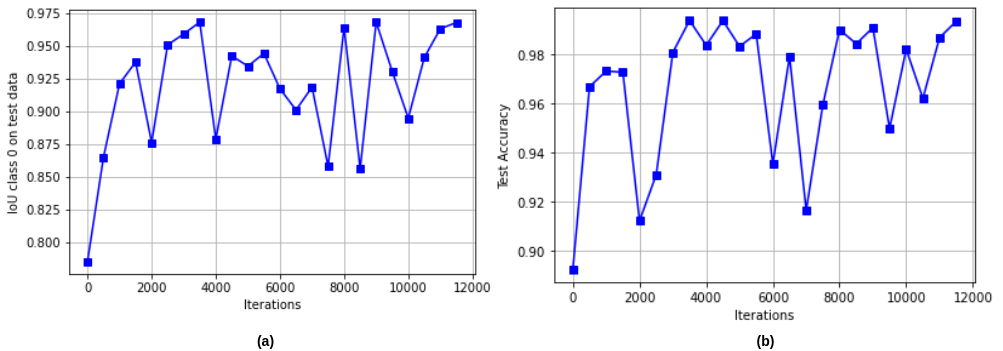} 
    \caption{Performance curves of $FC^2$MFN on test data. (a) Intersection over Union(IoU) of class 0. (b) Overall accuracy.}
    \label{fig:performance}
\end{figure*}

The segmentation performance of $FC^2$MFN is compared with other state-of-the-art networks. The input to real-valued networks is the amplitude of the master and slave SAR image and the interferometric phase angle.
For evaluating segmentation performance, overall accuracy and Intersection over Union are used.
Overall accuracy is defined as the ratio of correctly classified pixels and the total number of pixels.
Intersection over Union(IoU) is the ratio of intersection of ground truth labels with the predicted output and their union.
In Table \ref{tab:1}, the class-wise performance of $FC^2$MFN is compared with other segmentation networks by measuring the Intersection over Union(IoU) of all the three classes and their mean(Mean IoU).
$FC^2$MFN outperforms other networks for the classification of each class and the mean IoU.
Comparing $FC^2$MFN with CVCMFFNet, the IoU of shadow, ground and layover increases by \textbf{3.49\%}, \textbf{2.262\%} and \textbf{1.966\%} respectively. The Mean IoU increases by \textbf{31.46\%} compared to SegNet and by \textbf{4.318\%} compared to Deeplabv3+.
The Mean IoU increases by \textbf{2.5697\%} compared to CVCMFFNet.\\
Ablation study of complex-valued learning is given in Table \ref{tab:3}. Real, complex-valued, and fully complex-valued networks are compared based on overall accuracy(OA), mean pixel accuracy(MPA), and IoUs of different classes.
CV-SegNet, CV-SegNet+ASSP, CV-Deeplabv3+, and CVCMFFNet are termed complex-valued networks as the complex-valued data is projected in the real domain, and predictions are made using a real-valued output feature map.
$FC^2$MFN is a fully complex-valued network as the output is a complex-valued entity. Predictions are made using the real channel of the complex-valued output feature map.
Complex valued-networks (CV-networks) perform better than real-valued networks as shown in the table. Fully complex-valued network ($FC^2$MFN) outperforms both in all the metrics. The mean pixel accuracy (MPA) of $FC^2$MFN increases by \textbf{1.521\%} compared to CVCMFFNet.
Results from Table \ref{tab:1} and \ref{tab:3} suggest that retaining the complex-valued data throughout the network is a better approach for semantic segmentation of complex-valued InSAR input.\\
Overall test accuracy and IoU of class 0 of $FC^2$MFN is shown in Fig.\ref{fig:performance}.
$FC^2$MFN achieves high accuracy from the beginning without using any specific kind of pre-training or complex-valued weight initialization\cite{DBLP:journals/corr/TrabelsiBSSSMRB17} and has a smooth convergence. Similar results are observed for classes 1 and 2.
\begin{figure}[ht]
    \centering
    \includegraphics[width=8.55cm]{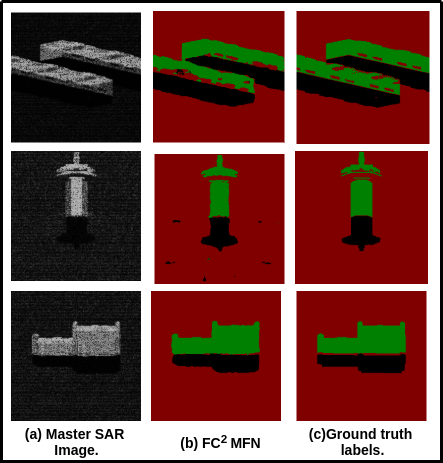}
    \caption{Image reconstructions of $FC^2$MFN are shown.The reconstructions are very similar to ground truth labels.}
    \label{fig:img_recon}
\end{figure}
The image reconstructions of $FC^2$MFN are shown in Fig.\ref{fig:img_recon}.
The master SAR image and ground truth label of some samples from the dataset are shown in Fig.\ref{fig:img_recon} (a,c).
The predictions made by $FC^2$MFN are shown in Fig.\ref{fig:img_recon} (b).These predictions are accurate and comparable to the ground truth labels.

\begin{table}
\caption{ Model evaluation of segmentation networks. \newline \textnormal{ }}
 \centering
 \setlength{\arrayrulewidth}{0.8pt}
\renewcommand{\arraystretch}{1.1}
\begin{tabular}{|c|c|c|c|c|}
\hline
Name       & Type                                                              & FLOPs           & \begin{tabular}[c]{@{}c@{}}Number of \\ parameters(M)\end{tabular} & \begin{tabular}[c]{@{}c@{}}Model size \\ (MB)\end{tabular} \\ \hline
UNet       & RV                                                                & 31.023M         & 31.033                                                             & 121.328                                                    \\
SegNet     & RV                                                                & 0.418T          & 29.429                                                             & 115.019                                                    \\
RefineNet  & RV                                                                & 1.461T          & 120.365                                                            & 470.723                                                    \\
PSPNet     & RV                                                                & 2.465M          & 2.408                                                              & 9.817                                                      \\
Deeplabv3+ & RV                                                                & 2.340T          & 118.782                                                            & 464.142                                                    \\
CVCMFFNet  & CV                                                                & 1.89T           & 84.150                                                             & 328.753                                                    \\ \hline
$FC^2$MFN   & \textbf{\begin{tabular}[c]{@{}c@{}}Fully \\ Complex\end{tabular}} & \textbf{1.82T} & \textbf{73.927}                                                    & 280                                                        \\ \hline
\multicolumn{4}{l}{The number of parameters are in millions (M).}
\end{tabular}
\label{tab:4}
\end{table}
 
\subsection{Model evaluation}
In Table \ref{tab:4}, the models are evaluated based on floating-point operations(FLOPs), number of parameters(in millions) and model size in megabytes(MB).
CVCMFFNet is a complex-valued network(CV), and $FC^2$MFN is a fully complex-valued network. All other networks are real-valued networks.
Although $FC^2$MFN has higher FLOPs and more parameters than UNet, SegNet, and PSPNet, as seen in table 4, its segmentation performance is much higher.
$FC^2$MFN has \textbf{3.227\%} lower FLOPs and \textbf{12.148\%} lesser number of parameters compared to CVCMFFNet.
This shows that $FC^2$MFN has a lower model complexity than CVCMFFNet and is more computationally efficient in terms of FLOPs.
\section{Conclusion}
In this paper, a fully complex-valued fully convolutional multi-feature fusion network for building semantic segmentation of InSAR images has been proposed. To deal with the particularity of complex-valued InSAR images, a novel, fully complex-valued learning scheme is used to train the network entirely in the complex domain without the need to project the data in the real domain using a complex-valued loss and complex-valued one-hot encoded labels.
To compare complex numbers in a fairer sense, complex-valued pooling has been proposed, which helps preserve the representation of complex numbers and helps the network learn better in the complex domain.
$FC^2$MFN performs multi-scale feature fusion and multi-channel feature fusion in a united framework and is evaluated on the simulated InSAR dataset\cite{2csm-3723-20}. Experiments show that the proposed network outperforms state-of-the-art networks like UNet, SegNet, RefineNet, DeepLabv3+, and CVCMFFNet.

\bibliographystyle{IEEEtran}
\bibliography{egbib}

\begin{thebibliography}{10}
\providecommand{\url}[1]{#1}
\csname url@samestyle\endcsname
\providecommand{\newblock}{\relax}
\providecommand{\bibinfo}[2]{#2}
\providecommand{\BIBentrySTDinterwordspacing}{\spaceskip=0pt\relax}
\providecommand{\BIBentryALTinterwordstretchfactor}{4}
\providecommand{\BIBentryALTinterwordspacing}{\spaceskip=\fontdimen2\font plus
\BIBentryALTinterwordstretchfactor\fontdimen3\font minus
  \fontdimen4\font\relax}
\providecommand{\BIBforeignlanguage}[2]{{%
\expandafter\ifx\csname l@#1\endcsname\relax
\typeout{** WARNING: IEEEtran.bst: No hyphenation pattern has been}%
\typeout{** loaded for the language `#1'. Using the pattern for}%
\typeout{** the default language instead.}%
\else
\language=\csname l@#1\endcsname
\fi
#2}}
\providecommand{\BIBdecl}{\relax}
\BIBdecl

\bibitem{2csm-3723-20}
\BIBentryALTinterwordspacing
J.~Chen, ``Simulated insar building dataset for cvcmff net,'' 2020. [Online].
  Available: \url{https://dx.doi.org/10.21227/2csm-3723}
\BIBentrySTDinterwordspacing

\bibitem{chen2021cvcmff}
J.~Chen, X.~Qiu, C.~Ding, and Y.~Wu, ``Cvcmff net: Complex-valued convolutional
  and multifeature fusion network for building semantic segmentation of insar
  images,'' \emph{IEEE Transactions on Geoscience and Remote Sensing}, vol.~60,
  pp. 1--14, 2021.

\bibitem{RAUCOULES2007289}
\BIBentryALTinterwordspacing
D.~Raucoules, C.~Colesanti, and C.~Carnec, ``Use of sar interferometry for
  detecting and assessing ground subsidence,'' \emph{Comptes Rendus
  Geoscience}, vol. 339, no.~5, pp. 289--302, 2007. [Online]. Available:
  \url{https://www.sciencedirect.com/science/article/pii/S1631071307000296}
\BIBentrySTDinterwordspacing

\bibitem{DBLP:journals/corr/RonnebergerFB15}
\BIBentryALTinterwordspacing
O.~Ronneberger, P.~Fischer, and T.~Brox, ``U-net: Convolutional networks for
  biomedical image segmentation,'' \emph{CoRR}, vol. abs/1505.04597, 2015.
  [Online]. Available: \url{http://arxiv.org/abs/1505.04597}
\BIBentrySTDinterwordspacing

\bibitem{zhao2017pyramid}
H.~Zhao, J.~Shi, X.~Qi, X.~Wang, and J.~Jia, ``Pyramid scene parsing network,''
  in \emph{Proceedings of the IEEE conference on computer vision and pattern
  recognition}, 2017, pp. 2881--2890.

\bibitem{lin2017refinenet}
G.~Lin, A.~Milan, C.~Shen, and I.~Reid, ``Refinenet: Multi-path refinement
  networks for high-resolution semantic segmentation,'' in \emph{Proceedings of
  the IEEE conference on computer vision and pattern recognition}, 2017, pp.
  1925--1934.

\bibitem{DBLP:journals/corr/abs-1802-02611}
\BIBentryALTinterwordspacing
L.~Chen, Y.~Zhu, G.~Papandreou, F.~Schroff, and H.~Adam, ``Encoder-decoder with
  atrous separable convolution for semantic image segmentation,'' \emph{CoRR},
  vol. abs/1802.02611, 2018. [Online]. Available:
  \url{http://arxiv.org/abs/1802.02611}
\BIBentrySTDinterwordspacing

\bibitem{Silberman:ECCV12}
P.~K. Nathan~Silberman, Derek~Hoiem and R.~Fergus, ``Indoor segmentation and
  support inference from rgbd images,'' in \emph{ECCV}, 2012.

\bibitem{Everingham15}
M.~Everingham, S.~M.~A. Eslami, L.~Van~Gool, C.~K.~I. Williams, J.~Winn, and
  A.~Zisserman, ``The pascal visual object classes challenge: A
  retrospective,'' \emph{International Journal of Computer Vision}, vol. 111,
  no.~1, pp. 98--136, Jan. 2015.

\bibitem{cordts2015cityscapes}
M.~Cordts, M.~Omran, S.~Ramos, T.~Scharw{\"a}chter, M.~Enzweiler, R.~Benenson,
  U.~Franke, S.~Roth, and B.~Schiele, ``The cityscapes dataset,'' in \emph{CVPR
  Workshop on the Future of Datasets in Vision}, vol.~2, 2015.

\bibitem{6789140}
T.~Nitta, ``Orthogonality of decision boundaries in complex-valued neural
  networks,'' \emph{Neural Computation}, vol.~16, no.~1, pp. 73--97, 2004.

\bibitem{9397870}
J.~Chen, X.~Qiu, C.~Ding, and Y.~Wu, ``Cvcmff net: Complex-valued convolutional
  and multifeature fusion network for building semantic segmentation of insar
  images,'' \emph{IEEE Transactions on Geoscience and Remote Sensing}, vol.~60,
  pp. 1--14, 2022.

\bibitem{long2015fully}
J.~Long, E.~Shelhamer, and T.~Darrell, ``Fully convolutional networks for
  semantic segmentation,'' 2015.

\bibitem{badrinarayanan2017segnet}
V.~Badrinarayanan, A.~Kendall, and R.~Cipolla, ``Segnet: A deep convolutional
  encoder-decoder architecture for image segmentation,'' \emph{IEEE
  transactions on pattern analysis and machine intelligence}, vol.~39, no.~12,
  pp. 2481--2495, 2017.

\bibitem{yu2022lightweight}
L.~Yu, Z.~Zeng, A.~Liu, X.~Xie, H.~Wang, F.~Xu, and W.~Hong, ``A lightweight
  complex-valued deeplabv3+ for semantic segmentation of polsar image,''
  \emph{IEEE Journal of Selected Topics in Applied Earth Observations and
  Remote Sensing}, 2022.

\bibitem{cao2019pixel}
Y.~Cao, Y.~Wu, P.~Zhang, W.~Liang, and M.~Li, ``Pixel-wise polsar image
  classification via a novel complex-valued deep fully convolutional network,''
  \emph{Remote Sensing}, vol.~11, no.~22, p. 2653, 2019.

\bibitem{DBLP:journals/corr/TrabelsiBSSSMRB17}
\BIBentryALTinterwordspacing
C.~Trabelsi, O.~Bilaniuk, D.~Serdyuk, S.~Subramanian, J.~F. Santos, S.~Mehri,
  N.~Rostamzadeh, Y.~Bengio, and C.~J. Pal, ``Deep complex networks,''
  \emph{CoRR}, vol. abs/1705.09792, 2017. [Online]. Available:
  \url{http://arxiv.org/abs/1705.09792}
\BIBentrySTDinterwordspacing

\bibitem{Simonyan15}
K.~Simonyan and A.~Zisserman, ``Very deep convolutional networks for
  large-scale image recognition,'' in \emph{International Conference on
  Learning Representations}, 2015.

\bibitem{lin2014network}
M.~Lin, Q.~Chen, and S.~Yan, ``Network in network,'' 2014.

\bibitem{Suresh2013}
S.~Suresh, N.~Sundararajan, and R.~Savitha, \emph{Circular Complex-valued
  Extreme Learning Machine Classifier}.\hskip 1em plus 0.5em minus 0.4em\relax
  Berlin, Heidelberg: Springer Berlin Heidelberg, 2013, pp. 109--123.

\bibitem{6414644}
R.~Savitha, S.~Suresh, and N.~Sundararajan, ``Projection-based fast learning
  fully complex-valued relaxation neural network,'' \emph{IEEE Transactions on
  Neural Networks and Learning Systems}, vol.~24, no.~4, pp. 529--541, 2013.

\bibitem{5236983}
R.~F.~H. Fischer, \emph{Appendix A: Wirtinger Calculus}, 2002, pp. 405--413.

\bibitem{Suresh2013-2}
S.~Suresh, N.~Sundararajan, and R.~Savitha, \emph{Fully Complex-valued Multi
  Layer Perceptron Networks}.\hskip 1em plus 0.5em minus 0.4em\relax Berlin,
  Heidelberg: Springer Berlin Heidelberg, 2013, pp. 31--47.

\bibitem{badrinarayanan2016segnet}
V.~Badrinarayanan, A.~Kendall, and R.~Cipolla, ``Segnet: A deep convolutional
  encoder-decoder architecture for image segmentation,'' 2016.

\bibitem{DBLP:journals/corr/KingmaB14}
\BIBentryALTinterwordspacing
D.~P. Kingma and J.~Ba, ``Adam: {A} method for stochastic optimization,'' in
  \emph{3rd International Conference on Learning Representations, {ICLR} 2015,
  San Diego, CA, USA, May 7-9, 2015, Conference Track Proceedings}, Y.~Bengio
  and Y.~LeCun, Eds., 2015. [Online]. Available:
  \url{http://arxiv.org/abs/1412.6980}
\BIBentrySTDinterwordspacing

\end{thebibliography}

\section*{A Appendix}
\subsection{Derivation of pooling layer.}
Let \textbf{C} and \textbf{R} denote the complex-valued domain and the real-valued domain respectively. 
Complex numbers $\emph{z}_1$ and $\emph{z}_2$ are represented as,\\
\begin{equation}
  \begin{aligned}
    z_1 & = x_1 + \emph{i}y_1 ,  \quad    \text{where} \quad  x_1,y_1 \in \textbf{R} \\ 
    z_2 & = x_2 + \emph{i}y_2 ,  \quad    \text{where} \quad  x_2,y_2 \in \textbf{R}
  \end{aligned}
\end{equation}

The complex-valued feature maps are converted from cartesian form to the polar form,
\begin{equation}
  \begin{aligned}
    z & = |r|\cdot e ^ {i\theta} = |r|\cdot(cos(\theta) + i \cdot sin(\theta)),
  \end{aligned}
\end{equation}
\begin{equation}
  \begin{aligned}
    \frac{1}{z} \ & = \frac{1}{|r|} \ e ^ {-i\theta} =  \frac{1}{|r|} \ \cdot(cos(\theta) - i \cdot sin(\theta)),
  \end{aligned}
\end{equation}
where  r = $\sqrt{x^2 + y^2}$  and $\theta=\arctan $ $ (\frac{y}{x})$. As the pooling layer is always applied after CRElu, the operation is always constrained to the first quadrant of the complex-plane,i.e, $\theta$ $\in$ [0,$\pi$/2].\\
To consider to the phase information. we take the square of magnitude of complex number z and its inverse,\\
\begin{equation}
  \begin{aligned}
    z^1 & = |z+ \frac{1}{z} |^2 \\
    & = | |r| \cdot(cos(\theta) + i \cdot sin(\theta)) + \frac{1}{|r|} \ \cdot(cos(\theta) - i \cdot sin(\theta)) |^2\\ 
     & =| |r| \cdot cos(\theta) + i \cdot |r| \cdot sin(\theta) + \frac{1}{|r|} \ cos(\theta) - i \cdot \frac{1}{|r|} \ sin(\theta) | ^2\\
     & = |(|r| + \frac{1}{|r|}\ ) \cdot cos(\theta) + i \cdot (|r| - \frac{1}{|r|}\ ) \cdot sin(\theta)| ^2\\
     & = (|r| \cdot cos(\theta) + \frac{1}{|r|}\ \cdot cos(\theta) ) ^2  + (|r| \cdot sin(\theta)  \\
     & - \frac{1}{|r|}\ \cdot sin(\theta) )^2 \\
     & = |r|^2 \cdot cos^2(\theta) + 2 \cdot cos^2(\theta) + \frac{1}{|r|^2}\ \cdot cos^2(\theta) \\ 
     &  + |r|^2 \cdot sin^2(\theta) - \quad 2 \cdot sin^2(\theta) +  \frac{1}{|r|^2}\ \cdot sin^2(\theta)  \\
     & = (|r|^2 + \frac{1}{|r|^2}\ ) \cdot cos^2(\theta)  + (|r|^2 + \frac{1}{|r|^2}\ ) \cdot sin^2(\theta)  \\
     & + 2 \cdot cos^2(\theta) - 2 \cdot sin^2(\theta) \\
     & = (|r|^2 + \frac{1}{|r|^2}\ )(cos^2(\theta) + sin^2(\theta)) \\
     & + 2 \cdot  (cos^2(\theta) - sin^2(\theta))\\
     & = (|r|^2 + \frac{1}{|r|^2}\ )(cos^2(\theta) + sin^2(\theta)) + 2 \cdot cos(2\theta)\\
      & = (|r|^2 + \frac{1}{|r|^2}\ ) + 2 \cdot cos(2\theta)
  \end{aligned}
\end{equation}
Eq. (11) is modified to (12) using hyperparameter $\delta$, mainly so that numbers with small radial components do not result in high likelihood of being chosen and for numerical stability.    
\begin{equation}
z^1 = \left\{
        \begin{array}{ll}
            |r_1|^2 + \frac{1}{|r_1|^2} + 2 \cdot cos(2 \theta_1), \quad |r_1| > \delta \\
            |r_1|^2  + 2 \cdot cos(2 \theta_1), \quad \quad \quad \text{otherwise.}
        \end{array}
    \right.
\end{equation}
Using (12), the feature maps are updated from z to $z^1$ and max-pooling operation is performed on the updated feature maps.

\end{document}